%% file: main.tex
\documentclass[10pt,twocolumn,letterpaper]{article}

\usepackage{cvpr}      
\usepackage{amsmath,amssymb}

\input{preamble}
\definecolor{cvprblue}{rgb}{0.21,0.49,0.74}
\usepackage[pagebackref,breaklinks,colorlinks,allcolors=cvprblue]{hyperref}

\def\paperID{*****} 
\def\confName{CVPR}
\def\confYear{2026}

\title{F3DGS: Federated 3D Gaussian Splatting for Decentralized Multi-Agent World Modeling}

\author{
Morui Zhu$^{1}$\thanks{Equal contribution.}\quad
Mohammad Dehghani Tezerjani$^{1}$\footnotemark[1]\\ \quad
Mátyás Szántó$^{2}$ \quad 
Márton Vaitkus$^{2}$\quad
Song Fu$^{1}$\quad
Qing Yang$^{1}$\\
$^{1}$University of North Texas\\
$^{2}$Budapest University of Technology and Economics\\
{\tt\small \{morui.zhu, mike.degany, song.fu, qing.yang\}@unt.edu}\\
{\tt\small \{szanto.matyas, vaitkus.marton\}@vik.bme.hu}
}

\begin{document}
\maketitle
\input{sec/0_abstract}    
\input{sec/1_intro}
\input{sec/2_related_work}
\input{sec/3_method}
\input{sec/4_experiments}
\input{sec/5_conclusion}
{
    \small
    \bibliographystyle{ieeenat_fullname}
    \bibliography{references}
}


\end{document}

%% file: sec/0_abstract.tex
\begin{abstract}
We present F3DGS, a federated 3D Gaussian Splatting framework for decentralized multi-agent 3D reconstruction.
Existing 3DGS pipelines assume centralized access to all observations, which limits their applicability in distributed robotic settings where agents operate independently, and centralized data aggregation may be restricted. Directly extending centralized training to multi-agent systems introduces communication overhead and geometric inconsistency.
F3DGS first constructs a shared geometric scaffold by registering locally merged LiDAR point clouds from multiple clients to initialize a global 3DGS model. During federated optimization, Gaussian positions are fixed to preserve geometric alignment, while each client updates only appearance-related attributes, including covariance, opacity, and spherical harmonic coefficients. The server aggregates these updates using visibility-aware aggregation, weighting each client's contribution by how frequently it observed each Gaussian, resolving the partial-observability challenge inherent to multi-agent exploration. 
To evaluate decentralized reconstruction, we collect a multi-sequence indoor dataset with synchronized LiDAR, RGB, and IMU measurements. Experiments show that F3DGS achieves reconstruction quality comparable to centralized training while enabling distributed optimization across agents. 
The dataset, development kit, and source code will be publicly released.
\end{abstract}

%% file: sec/1_intro.tex
\section{Introduction}
\label{sec:intro}

3D Gaussian Splatting (3DGS)~\cite{kerbl3Dgaussians} represents scenes as collections of anisotropic 3D Gaussian primitives rendered via tile-based differentiable rasterization, achieving state-of-the-art novel-view synthesis at interactive frame rates. This combination of quality and efficiency has driven adoption across robotics~\cite{matsuki2024gaussian}, autonomous driving~\cite{yan2024street}, and embodied AI~\cite{wu2024recent}.

All current 3DGS methods, however, share a restrictive assumption: \textit{centralized access to the complete set of training observations}. That is, every image from every viewpoint must reside on a single machine for joint optimization. This assumption holds in controlled single-agent capture but collapses in distributed robotic settings where multiple agents independently collect observations and centralized data aggregation is impractical or prohibited.

Realizing 3DGS in multi-agent systems requires reconciling three constraints that centralized pipelines fundamentally cannot satisfy. \textbf{(1)~Communication cost:} aggregating high-resolution images from multiple agents at a centralized server incurs bandwidth and storage costs that scale linearly with the number of agents, rapidly exceeding the practical limits of field-deployable infrastructure. \textbf{(2)~Data privacy:} in multi-operator or multi-organization settings, each agent’s sensor data is private and cannot be directly shared or combined unless data-sharing policies are satisfied, which is complex to enforce in distributed systems. \textbf{(3)~Scalability:} joint optimization over the union of all agent observations couples memory and compute requirements to the total fleet size, creating a centralized bottleneck that grows with deployment scale.

Federated learning~\cite{mcmahan2017communication} offers a promising direction, wherein agents keep data local and share only model updates with a coordinating server.
However, applying standard federated averaging to 3DGS introduces two domain-specific challenges.
\textbf{Geometric drift.} Unlike neural network weights that occupy a smooth parameter space, Gaussian primitives encode explicit 3D positions. Independent optimization across clients causes positions to drift into geometrically inconsistent configurations. Averaging drifted positions yields an incoherent reconstruction.
\textbf{Partial observability.} Each client observes only a contiguous temporal segment. Many Gaussians are visible to some clients but not others. Uniform averaging dilutes well-observed appearance estimates with values from non-observing clients.

To tackle these challenges, we present \textbf{F3DGS}, a federated 3DGS framework that addresses both challenges through geometry-appearance decoupling. The approach proceeds in three stages. First, a shared geometric scaffold is constructed by fusing LiDAR point clouds from all clients into a common world frame via metric-scale pose stitching. This scaffold initializes Gaussian primitives at fixed positions. Second, during federated optimization, all positions are frozen; each client optimizes only appearance attributes: scale, rotation, opacity, and spherical harmonic coefficients, on its local observations. This eliminates geometric drift by construction. Third, the server aggregates updates via \textit{visibility-aware weighting}: each client's influence on a given Gaussian is proportional to how frequently that Gaussian was rendered during local training.
For globally consistent agent trajectories, we align agent-level camera poses to a global pointcloud map with smooth boundary interpolation, producing metric-scale global trajectories without centralized bundle adjustment.


The contributions of our paper are summarized as follows:
\begin{itemize}
    \item A federated 3DGS framework that decouples a shared frozen geometric scaffold from locally optimized, visibility-weighted appearance attributes, enabling distributed reconstruction without raw data exchange.
    \item A trajectory stitching pipeline aligning agent-level monocular trajectories to a LiDAR SLAM anchor with boundary smoothing for globally consistent metric-scale poses.
    \item Empirical validation on nine indoor sequences showing that federated aggregation preserves reconstruction quality within 2\,dB of locally trained models.
\end{itemize}

%% file: sec/2_related_work.tex
\section{Related Work}
\label{sec:related_work}

\subsection{3D Gaussian Splatting}

Kerbl~\etal~\cite{kerbl3Dgaussians} introduced 3DGS, optimizing anisotropic Gaussian primitives through differentiable rasterization with adaptive density control. Extensions include 2DGS~\cite{huang20242d} for surface-aligned primitives, Mip-Splatting~\cite{yu2024mip} for multi-scale anti-aliasing, and several SLAM integrations: GS-SLAM~\cite{yan2024gs} for real-time dense mapping, SplaTAM~\cite{keetha2024splatam} for silhouette-guided densification, and Photo-SLAM~\cite{huang2024photo} for feature-based tracking with Gaussian mapping, ReLoc-3DGS~\cite{ye2025reloc} replacing SfM with LiDAR odometry for initialization.

All above methods require centralized access to the full observation set. To our knowledge, exisiting 3DGS methods do not address settings where observations are distributed across agents that cannot share raw data. F3DGS formulates 3DGS under an explicit federated constraint.

\subsection{Federated Learning}

FedAvg~\cite{mcmahan2017communication} established the broadcast-train-aggregate paradigm for distributed model optimization. Subsequent work addresses client heterogeneity through variance reduction~\cite{karimireddy2020scaffold}, personalization layers~\cite{arivazhagan2019federated}, and proximal regularization~\cite{li2020federated}. Applications in vision span classification~\cite{mcmahan2017communication}, detection~\cite{chen2020fedbe}, and segmentation~\cite{fantauzzo2022feddrive}.

FedNeRF~\cite{suzuki2024fednerf} applied federated averaging to neural radiance fields by aggregating MLP weights. However, NeRF's implicit representation entangles geometry and appearance in shared network parameters, precluding selective freezing. 3DGS stores positions, covariances, and colors as explicit, separable tensors, a structure we exploit by federating appearance while fixing geometry. Fed3DGS~\cite{suzuki2024fed3dgs} extends federated learning to 3DGS via distillation-based server-side model updates with appearance modeling, but does not prevent geometric drift across clients and incurs additional rendering cost at each aggregation step.

\subsection{Multi-Agent 3D Reconstruction}

Collaborative SLAM methods exchange submaps~\cite{lajoie2024swarm}, keyframes~\cite{schmuck2019ccm}, learned descriptors~\cite{tian2022kimera}, or localized scans~\cite{dehghani2025networking} to build shared geometric maps. These produce point clouds or occupancy grids but not photorealistic renderable representations. Block-NeRF~\cite{tancik2022block} partitions large scenes into independently trained NeRFs merged through appearance alignment, but operates under centralized data access per block.
CoSurfGS~\cite{gao2024cosurfgs} enables distributed Gaussian surface reconstruction through a device-edge-cloud hierarchy with self-distillation-based merging, but assumes cooperative model sharing rather than enforcing a federated data constraint.

F3DGS differs on two axes: it enforces the federated constraint (zero raw image exchange), and it produces a single unified Gaussian model through iterative aggregation rather than post-hoc merging of independent submaps.

\subsection{LiDAR-Visual Pose Estimation}

Accurate metric-scale pose estimation is a prerequisite for any multi-agent reconstruction pipeline, and a wide range of LiDAR and visual odometry methods exist for this purpose. KISS-ICP~\cite{vizzo2023kiss} and KISS-SLAM~\cite{guadagnino2025kiss} provide robust LiDAR odometry and SLAM with loop closure, while Depth Anything~\cite{yang2024depth} predicts dense metric-scale depth from monocular images, enabling camera pose estimation without LiDAR. Our pose construction is backend-agnostic; we demonstrate both KISS-SLAM and Depth Anything integration, using Umeyama~\cite{umeyama1991least} alignment to stitch agent-level trajectories into globally consistent metric-scale poses.

%% file: sec/3_method.tex
\section{Method}

\begin{figure*}
    \centering
    \includegraphics[width=1.0\linewidth]{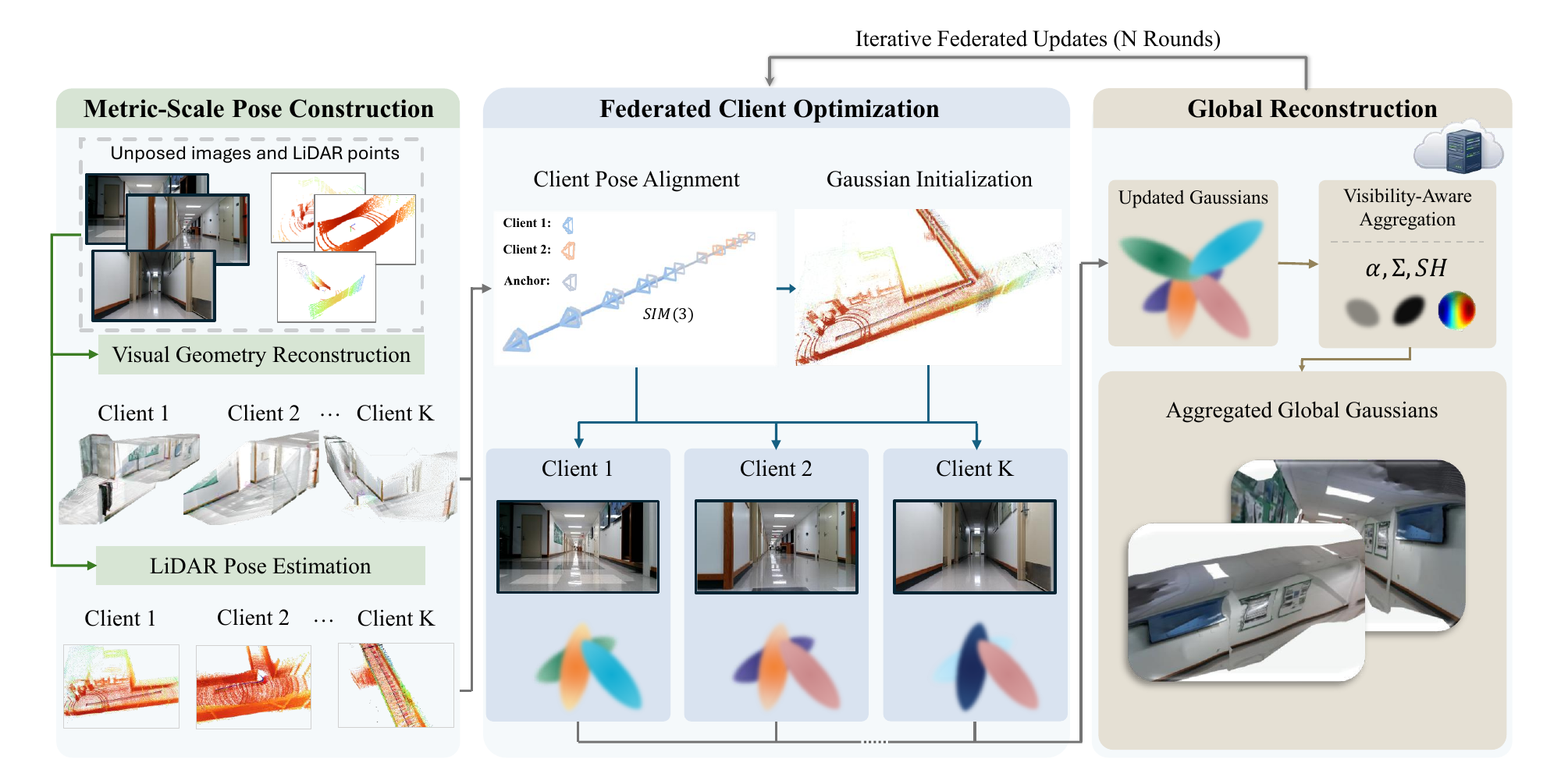}
    \caption{Overview of our federated 3D Gaussian splatting (F3DGS) framework for learning consistent global scene representations. Each client first constructs metric-scale poses from unposed images and LiDAR via visual geometry reconstruction and LiDAR odometry. Based on these poses, clients perform local optimization, including pose alignment and Gaussian initialization, to obtain per-client 3D Gaussian representations. Through iterative federated updates, client models are progressively aggregated. Finally, a visibility-aware fusion strategy integrates Gaussian attributes (e.g., opacity, scale, and spherical harmonics) to produce a unified and consistent global reconstruction.}
    \label{fig:pipeline}
\end{figure*}

\subsection{Overview}

We study the learning of 3D Gaussian scene representations from a federated learning perspective. 
In many real-world scenarios, visual data are naturally distributed across multiple devices or agents. 
Instead of centralizing raw images for training, federated learning supports collaborative model optimization while keeping data local to each client.

In this work, we formulate 3D Gaussian Splatting (3DGS) training under a federated framework where each client optimizes a local subset of observations while contributing to a shared global Gaussian representation through iterative aggregation. The framework, as illustrated in Fig.~\ref{fig:pipeline}, consists of three main components:
(1) metric-scale camera pose construction using modern pose estimation pipelines,
(2) federated optimization of Gaussian attributes with frozen geometry for stable distributed training, and
(3) visibility-aware aggregation that integrates updates from multiple clients.
%

\subsection{Problem Formulation}

A scene is observed through $N$ images (RGB) with camera poses $\mathcal{D} = \{(I_t, T_t)\}_{t=1}^{N}$, where $I_t$ denotes the image captured at time $t$ and $T_t \in \text{SE}(3)$ is the camera-to-world transformation. The scene is represented by $M$ Gaussian primitives:
\begin{equation}
    \mathcal{G} = \{g_i\}_{i=1}^{M}, \quad g_i = (\mu_i, s_i, q_i, \alpha_i, \mathbf{c}_i),
    \label{eq:gaussians}
\end{equation}
where $\mu_i \in \mathbb{R}^3$ is the center, $s_i \in \mathbb{R}^3$ the log-scale, $q_i \in \mathbb{R}^4$ the orientation quaternion, $\alpha_i \in \mathbb{R}$ the logit-opacity, and $\mathbf{c}_i$ the spherical harmonic (SH) color coefficients.

To enable federated training, the full sequence is partitioned into $K$ clients.
Each client $k$ contains a subset of frames of size $C$,

\begin{equation}
\mathcal{D}_k = \{(I_t, T_t) \mid t \in \mathcal{I}_k\},
\end{equation}
where $\mathcal{I}_k = [kC,\, \min((k{+}1)C{-}1,\, N{-}1)]$. 
The objective is to learn a shared Gaussian representation by minimizing

\begin{equation}
    \min_{\mathcal{G}} \sum_{k=1}^{K} \mathcal{L}_k(\mathcal{G}),
    \label{objective}
\end{equation}
where $\mathcal{L}_k$ (defined in~ \ref{eq:loss}) denotes the rendering loss evaluated on client $k$.

\subsection{Metric-Scale Pose Construction}

Camera trajectories are estimated using modern monocular pose estimation methods that provide metric-scale reconstruction.
It is important to note that our framework does not rely on a specific pose estimation algorithm; instead, any pose estimation pipeline that produces metric-scale trajectories can be used.

Geometric consistency across clients requires globally aligned metric-scale poses. Let $T_t^{(k)}$ denote poses predicted within client $k$.
We align each client trajectory to a global anchor, achieved by a global LiDAR-based trajectory estimation, via Umeyama $\mathcal{S}im(3)$ estimation~\cite{umeyama1991least}, computing scale $s_k$, rotation $R_k$, and translation $\mathbf{t}_k$:

\begin{equation}
    s_k, R_k, \mathbf{t}_k = \arg\min_{s, R, \mathbf{t}} \sum_{j} \| \mathbf{p}_j^{\text{anchor}} - (s R \mathbf{p}_j^{\text{client}} + \mathbf{t}) \|^2.
    \label{eq:sim3}
\end{equation}

To prevent boundary discontinuities, we apply exponentially decayed $\text{SE(3)}$ corrections over a smoothing window of $\tau$ frames:
\begin{equation}
    T_t^{\text{smooth}} = \text{Exp}\!\left(\beta(t) \cdot \text{Log}(\Delta T)\right) \cdot T_t^{\text{aligned}},
    \label{eq:smooth}
\end{equation}
where $\beta(t) = 1 - t/\tau$ linearly decays the boundary correction $\Delta T$ to identity.
This procedure produces a globally consistent pose sequence that is used for Gaussian optimization.








\subsection{Global Gaussian Initialization}
Before federated optimization, we initialize a global 3D Gaussian representation of the scene. 
We first project depth observations into 3D space to obtain dense point samples, which are accumulated across frames to form an initial point cloud. 
Each point is then converted into a Gaussian primitive with its mean centered at the point location. 
Color attributes are initialized from the corresponding RGB observations, while scales are set from KNN distance to 3 nearest neighbors, and opacities as small positive values.


In our implementation, we initialize approximately $6\times10^5$ Gaussian primitives to ensure sufficient coverage of the scene geometry.

\subsection{Local Federated Optimization}

Federated training proceeds in multiple communication rounds.
Each round, the server broadcasts global parameters to all clients. 
Each client performs local optimization using only its local image subset.
The rendering loss for client $k$ combines L1 and SSIM:
\begin{equation}
    \mathcal{L}_k = \sum_{t \in \mathcal{I}_k} \left[ (1{-}\lambda)\, \| I_t - \hat{I}_t \|_1 + \lambda\, (1 - \text{SSIM}(I_t, \hat{I}_t)) \right],
    \label{eq:loss}
\end{equation}
with $\lambda = 0.2$, where $\hat{I}_t$ denotes the rendered image generated from the Gaussian representation.
%

\paragraph{Frozen geometry.} Gaussian positions are fixed throughout local training:
\begin{equation}
    \mu_i^{(k)} = \mu_i \quad \forall\, k, \; \forall\, \text{steps}.
    \label{eq:freeze}
\end{equation}
Only appearance parameters $\theta_{\text{app}} = \{s_i, q_i, \alpha_i, \mathbf{c_i}\}_{i=1}^{M}$ receive gradient updates.
%
%
%
This design prevents geometric drift across clients and improves the stability of federated optimization.

\paragraph{Visibility tracking.} A per-Gaussian counter $v_{k,i}$ accumulates the number of rasterization passes in which Gaussian $i$ had positive projected radius during client $k$'s training.

\subsection{Visibility-Aware Federated Aggregation}
\label{sec:aggregation}

Each client transmits updated appearance attributes $\theta_{\text{app}}^{(k)}$ and visibility counts $\{v_{k,i}\}$ to the server.

\paragraph{Aggregation weights.} For Gaussian $i$, client $k$'s weight is:
\begin{equation}
    \alpha_{k,i} = \frac{v_{k,i}}{\sum_{j=1}^{K} v_{j,i} + \epsilon}, \quad \epsilon = 10^{-8}.
    \label{eq:vis_weight}
\end{equation}
The global attribute is the weighted combination:
\begin{equation}
    a_i = \sum_{k=1}^{K} \alpha_{k,i} \, a_{k,i}.
    \label{eq:agg}
\end{equation}
Gaussians with zero total visibility retain their previous global values.

Quaternion parameters require special handling due to sign ambiguity.
We align quaternion signs before averaging and normalize the resulting quaternion to ensure valid rotations.

%% file: sec/4_experiments.tex
\section{Experiments}

\begin{table*}[t]
\centering
\small
\setlength{\tabcolsep}{6pt}
\begin{tabular}{c c c c c ccc ccc}
\toprule
Seq & Frames & Clients & Split & Val &
\multicolumn{3}{c}{Local} &
\multicolumn{3}{c}{Global} \\
\cmidrule(lr){6-8} \cmidrule(lr){9-11}
& $N$ & $K$ & & &
PSNR$\uparrow$ & SSIM$\uparrow$ & LPIPS$\downarrow$ &
PSNR$\uparrow$ & SSIM$\uparrow$ & LPIPS$\downarrow$ \\
\midrule
03 & 1840 & 5 & 400/400/400/400/240 & 230 & 22.36 & 0.7826 & 0.4284 & 18.82 & 0.7111 & 0.5234 \\
04 & 1407 & 4 & 400/400/400/207     & 176 & 22.96 & 0.7889 & 0.4016 & 20.50 & 0.7431 & 0.4684 \\
05 & 543  & 2 & 400/143             & 68  & 23.71 & 0.8182 & 0.3910 & 22.65 & 0.8076 & 0.4154 \\
06 & 1042 & 3 & 400/400/242         & 131 & 25.66 & 0.8310 & 0.3613 & 22.66 & 0.8027 & 0.4104 \\
07 & 763  & 2 & 400/363             & 96  & 23.64 & 0.8113 & 0.4049 & 22.74 & 0.8005 & 0.4227 \\
08 & 718  & 2 & 400/318             & 90  & 24.52 & 0.8530 & 0.3741 & 23.94 & 0.8315 & 0.3882 \\
09 & 1018 & 3 & 400/400/218         & 128 & 25.03 & 0.8548 & 0.3418 & 21.85 & 0.8260 & 0.3931 \\
10 & 856  & 3 & 400/400/56          & 107 & 25.07 & 0.8449 & 0.3525 & 22.87 & 0.8118 & 0.3920 \\
11 & 552  & 2 & 400/152             & 69  & 24.01 & 0.8274 & 0.3664 & 22.77 & 0.8078 & 0.3853 \\
\bottomrule
\end{tabular}
\caption{Sequence-level quantitative results on meangreen dataset sequences 03--11. Local metrics evaluate each client model on its own validation split, while global metrics evaluate the final aggregated server model on the union of all validation splits. All sequence-level values are weighted by the number of validation images.}
\label{tab:main_results}
\end{table*}

\subsection{Setup}
Experiments are conducted on the MeanGreen dataset, which consists of long indoor sequences collected by a mobile robot with a monocular camera and a Velodyne LiDAR.
Each sequence is treated as a distributed observation stream and partitioned into temporally contiguous clients with chunk size $C{=}400$. 
We evaluate on sequences 03--11, resulting in varying numbers of clients per sequence.

A single global Gaussian representation is initialized per sequence using fused observations and shared across all clients. 
The number of primitives is fixed at $6\times10^5$.
Federated training is performed for $R{=}7$ rounds with $T{=}1000$ local optimization steps per round. 
Gaussian centers are kept fixed throughout training, and only appearance parameters are updated. 
Adaptive density control and any form of post-aggregation alignment are disabled.

\subsection{MeanGreen Dataset}

\begin{table}[t]
\centering
\small
\setlength{\tabcolsep}{6pt}
\begin{tabular}{l l}
\toprule
Sensor & Details \\
\midrule
1$\times$ Camera &
RGB (OAK-D-Pro), $1280\times720$, 63° FOV \\
1$\times$ LiDAR &
Velodyne VLP-16, 16 channels, 10\,Hz rotation, \\
& 100\,m range, $-15^\circ$ to $15^\circ$ vertical FOV \\
\bottomrule
\end{tabular}
\caption{Sensor specifications of the MeanGreen dataset.}
\label{tab:dataset_sensor}
\end{table}

We evaluate our work on part of the MeanGreen dataset, which contains 9 indoor corridor sequences (03--11) with a total of 8,739 frames. 
The data are collected using a mobile platform equipped with an RGB camera and a LiDAR sensor (Table~\ref{tab:dataset_sensor}). The camera intrinsics are calibrated, and camera--LiDAR extrinsics are obtained using a standard LiDAR--camera calibration tool~\cite{calibration}. 
These calibrations are used for pose construction and local federated optimization. The sequences are recorded in indoor corridor environments, including straight paths and turning motions. 
The scenes exhibit sparse texture, repeated structures, and a forward-facing camera with limited field of view, which restricts viewpoint diversity.

The MeanGreen dataset is currently a work in progress. A more comprehensive version, including both indoor and outdoor sequences, will be introduced in a forthcoming publication.

\subsection{Evaluation Protocol}

We report Peak Signal-to-Noise Ratio (PSNR) ($\uparrow$), Structural Similarity Index Measure (SSIM) ($\uparrow$), and Learned Perceptual Image Patch Similarity (LPIPS) ($\downarrow$) for all experiments.
Two evaluation protocols are used. 
Local evaluation measures reconstruction quality on each client's validation frames (every 8th frame), reflecting within-client fitting. 
Global evaluation measures the aggregated model on the union of all validation frames, reflecting cross-client consistency after federated aggregation.
Sequence-level metrics are computed as weighted averages with respect to the number of validation images.

\subsection{Quantitative Results}

Table~\ref{tab:main_results} reports sequence-level results on sequences 03--11. 
Local reconstruction quality is consistently high across all sequences, with PSNR exceeding 22 in all cases. After aggregation, global performance remains close to local performance on most sequences. The gap between local and global metrics is small for sequences such as 08, 10, and 11, indicating that the shared representation can be maintained under distributed optimization. The main degradation occurs on sequence 03, where global PSNR drops to 18.82. This indicates that aggregation remains sensitive to inconsistencies introduced across temporal partitions. 
Sequence 09 exhibits a similar but less severe trend.

Overall, the results show that the performance drop from local to global models is not uniform across sequences, and depends on how well observations from different clients are aligned in the shared representation.

\begin{table*}[t!]
\centering
\begin{tabular*}{0.8\linewidth}{@{\extracolsep{\fill}} c c c ccc ccc @{}}
\toprule
Seq & $R$ & $T$ &
\multicolumn{3}{c}{Local} &
\multicolumn{3}{c}{Global} \\
\cmidrule(lr){4-6} \cmidrule(lr){7-9}
& & &
PSNR$\uparrow$ & SSIM$\uparrow$ & LPIPS$\downarrow$ &
PSNR$\uparrow$ & SSIM$\uparrow$ & LPIPS$\downarrow$ \\
\midrule
07 & 1  & 7000 & 23.899 & 0.8137 & 0.3988 & 21.952 & 0.7918 & 0.4456 \\
07 & 2  & 3500 & 23.839 & 0.8126 & 0.4003 & 22.349 & 0.7962 & 0.4304 \\
07 & 5  & 1400 & 23.710 & 0.8116 & 0.4031 & 22.628 & 0.7993 & 0.4235 \\
07 & 7  & 1000 & 23.644 & 0.8113 & 0.4049 & 22.740 & 0.8005 & 0.4227 \\
07 & 14 & 500  & 23.568 & 0.8103 & 0.4056 & 22.837 & 0.8021 & 0.4194 \\
\midrule
08 & 1  & 7000 & 25.176 & 0.8589 & 0.3648 & 23.689 & 0.8324 & 0.3931 \\
08 & 2  & 3500 & 24.801 & 0.8558 & 0.3701 & 23.951 & 0.8355 & 0.3871 \\
08 & 5  & 1400 & 24.654 & 0.8545 & 0.3717 & 23.982 & 0.8361 & 0.3868 \\
08 & 7  & 1000 & 24.520 & 0.8530 & 0.3741 & 23.940 & 0.8315 & 0.3882 \\
08 & 14 & 500  & 24.382 & 0.8515 & 0.3791 & 24.009 & 0.8374 & 0.3876 \\
\bottomrule
\end{tabular*}
\caption{Effect of communication frequency under a fixed training budget of 7000 steps.}
\label{tab:schedule_ablation}
\end{table*}


\subsection{Client-Level Analysis}

To localize the source of global degradation, we further examine reconstruction quality at the client level, as shown in Table~\ref{tab:per_client_results}. Performance drop is concentrated in a small subset of clients rather than distributed across all partitions. For example, in sequence 03, client1 drops from 19.97 local PSNR to 15.44 after aggregation, and in sequence 09, client2 drops from 29.28 to 19.53. These clients dominate the sequence-level degradation observed in Table~\ref{tab:main_results}.

In contrast, many clients show minimal discrepancy between local and global performance. Examples include sequence 08 client0 (25.66 vs.\ 25.61) and sequence 10 client0 (26.98 vs.\ 26.89), where aggregation preserves local reconstruction quality.

These results indicate that global performance is primarily limited by a small number of difficult temporal segments, rather than a systematic failure across all clients.

\begin{table}[t]
\centering
\small
\setlength{\tabcolsep}{4pt}
\begin{tabular}{c ccc ccc}
\toprule
Client &
\multicolumn{3}{c}{Local} &
\multicolumn{3}{c}{Global} \\
\cmidrule(lr){2-4} \cmidrule(lr){5-7}
&
PSNR$\uparrow$ & SSIM$\uparrow$ & LPIPS$\downarrow$ &
PSNR$\uparrow$ & SSIM$\uparrow$ & LPIPS$\downarrow$ \\
\midrule
03-0 & 24.11 & 0.8143 & 0.3811 & 21.10 & 0.7758 & 0.4235 \\
03-1 & 19.97 & 0.7438 & 0.5035 & 15.44 & 0.6213 & 0.6383 \\
03-2 & 22.97 & 0.7922 & 0.4052 & 19.05 & 0.7145 & 0.5300 \\
03-3 & 22.26 & 0.7759 & 0.4333 & 19.28 & 0.7146 & 0.5201 \\
03-4 & 22.56 & 0.7896 & 0.4123 & 19.47 & 0.7416 & 0.4927 \\
04-0 & 24.34 & 0.8350 & 0.3740 & 23.28 & 0.8164 & 0.3950 \\
04-1 & 22.18 & 0.7732 & 0.4085 & 18.83 & 0.7019 & 0.5159 \\
04-2 & 23.29 & 0.7908 & 0.4072 & 19.82 & 0.7308 & 0.4948 \\
04-3 & 21.17 & 0.7265 & 0.4307 & 19.70 & 0.7052 & 0.4676 \\
05-0 & 23.18 & 0.7995 & 0.3946 & 23.02 & 0.7969 & 0.3973 \\
05-1 & 25.19 & 0.8703 & 0.3812 & 21.61 & 0.8374 & 0.4658 \\
06-0 & 23.71 & 0.7879 & 0.3992 & 20.40 & 0.7552 & 0.4318 \\
06-1 & 25.32 & 0.8240 & 0.3770 & 23.11 & 0.7938 & 0.4423 \\
06-2 & 29.34 & 0.9116 & 0.2750 & 25.59 & 0.8936 & 0.3243 \\
07-0 & 25.30 & 0.8281 & 0.3736 & 24.36 & 0.8188 & 0.3842 \\
07-1 & 21.84 & 0.7930 & 0.4388 & 20.98 & 0.7806 & 0.4646 \\
08-0 & 25.66 & 0.8457 & 0.3637 & 25.61 & 0.8450 & 0.3641 \\
08-1 & 23.10 & 0.8623 & 0.3870 & 21.85 & 0.8145 & 0.4183 \\
09-0 & 22.70 & 0.8164 & 0.3681 & 21.45 & 0.7954 & 0.3971 \\
09-1 & 24.98 & 0.8662 & 0.3624 & 23.54 & 0.8558 & 0.3853 \\
09-2 & 29.28 & 0.9030 & 0.2583 & 19.53 & 0.8272 & 0.4001 \\
10-0 & 26.98 & 0.8818 & 0.3184 & 26.89 & 0.8809 & 0.3193 \\
10-1 & 23.36 & 0.8200 & 0.3897 & 19.23 & 0.7622 & 0.4636 \\
10-2 & 23.77 & 0.7595 & 0.3301 & 20.17 & 0.6721 & 0.3997 \\
11-0 & 24.93 & 0.8642 & 0.3444 & 24.23 & 0.8522 & 0.3528 \\
11-1 & 21.59 & 0.7308 & 0.4240 & 18.92 & 0.6910 & 0.4706 \\
\bottomrule
\end{tabular}
\caption{Per-client local and global reconstruction metrics across sequences 03--11.}
\label{tab:per_client_results}
\end{table}

\subsection{Effect of Communication Frequency}

We vary the number of communication rounds $R$ and local steps $T$ under a fixed total budget of 7000 optimization steps.
Local reconstruction remains stable across all configurations. 
On both sequences, local PSNR changes by less than 0.4\,dB as $R$ increases.

In contrast, global performance improves steadily with more frequent aggregation. 
On sequence 07, global PSNR increases from 21.95 at $R{=}1$ to 22.74 at $R{=}7$, with a smaller gain thereafter (22.84 at $R{=}14$). 
A similar pattern appears on sequence 08.
The improvement is most pronounced in the low-round regime, while additional rounds yield progressively smaller gains. 
Beyond $R{=}7$, the increase becomes marginal relative to the added communication cost.
We therefore adopt $R{=}7$ and $T{=}1000$ in all experiments as a balanced configuration.


%
\subsection{Discussion and Limitations}

Two consistent observations emerge from the experiments. 
First, local optimization is stable across all sequences, indicating that fixing Gaussian geometry does not hinder within-client reconstruction. Second, global degradation is concentrated near difficult temporal boundaries and short segments, rather than being uniformly distributed.

The current formulation has two main limitations. Since Gaussian centers are fixed, the model cannot correct residual pose errors after initialization. In addition, performance remains sensitive to the choice of partitioning, especially in sequences with large appearance or motion changes across client boundaries.

%% file: sec/5_conclusion.tex
\section{Conclusions}

We introduced F3DGS, a federated 3D Gaussian Splatting framework for decentralized multi-agent 3D reconstruction. Our key idea is to decouple geometry from appearance: a shared geometric scaffold is initialized from registered LiDAR point clouds, while federated training updates only appearance-related Gaussian parameters on each client. Combined with visibility-aware aggregation, this design addresses two central challenges in distributed 3DGS: geometric drift and partial observability. Experiments on the MeanGreen dataset demonstrate that F3DGS achieves strong local reconstruction quality and, in most cases, maintains competitive global performance after aggregation, supporting the feasibility of federated scene reconstruction without raw data sharing.

Several directions remain for future work. Improving robustness under difficult temporal partitioning and at client boundaries is an important next step, as these cases currently account for most of the observed global degradation. Further evaluation against related baselines will also help position the method more clearly within the broader literature on decentralized reconstruction. In addition, we plan to extend the study to more challenging indoor and outdoor datasets and to release the MeanGreen dataset, code, and development tools to facilitate future work in this area.